\documentclass[10pt,twocolumn,letterpaper]{article}

\usepackage{graphicx}
\usepackage{subcaption}
\usepackage{float}
\usepackage[justification=raggedright]{caption}	%
\usepackage{lscape}                                         %
\usepackage{wrapfig}

\usepackage[lined,ruled,linesnumbered]{algorithm2e}
\usepackage{animate} %

\usepackage{booktabs}                   %
\usepackage{multirow}

\usepackage{makecell}

\usepackage{paralist}
\usepackage{enumitem}
\setlist[enumerate]{itemsep=0mm}

\usepackage{bm}                          %
\usepackage{epsfig}                      %
\usepackage{graphicx}                  %
\usepackage{times}
\usepackage{mathptmx}
\usepackage{mathtools}
\usepackage{amssymb,amsmath}   %

\usepackage{units}
\usepackage{color}

\usepackage[pagebackref,breaklinks,colorlinks,linkcolor=blue,citecolor=blue,urlcolor=blue,bookmarks=false]{hyperref}
\usepackage{xspace}
\usepackage[table]{xcolor}
\usepackage{setspace}
\usepackage{grfext}
\PrependGraphicsExtensions*{.jpg,.png,.PNG}

\usepackage{gensymb}

\usepackage{soul}
\usepackage{svg}

\usepackage[accsupp]{axessibility} %

\usepackage[final]{changes}

\def\etal{et~al.\_}			  %

\DeclareMathAlphabet{\altmathcal}{OMS}{cmsy}{m}{n}
\DeclareMathAlphabet{\mathbfit}{OT1}{ptm}{bx}{it}

\newlength\paramargin
\newlength\figmargin

\newlength\secmargin
\newlength\figcapmargin
\newlength\tabcapmargin

\newcommand{\mpage}[2]
{
\begin{minipage}{#1\linewidth}\centering
#2
\end{minipage}
}

\newcommand{\mfigure}[2]
{
\includegraphics[width=#1\linewidth]{#2}
}

\newcommand{\topic}[1]
{
\vspace{1mm}\noindent\textbf{#1}
}

\newcommand{\secref}[1]{Section~\ref{sec:#1}}

\long\def\ignorethis#1{}

\newbox\jsavebox%

\makeatletter
\newcommand{\providelength}[1]{%
  \@ifundefined{\expandafter\@gobble\string#1}
   {%
    \typeout{\string\providelength: making new length \string#1}%
    \newlength{#1}%
   }
   {%
    \sdaau@checkforlength{#1}%
   }%
}

\newcommand{\sdaau@checkforlength}[1]{%
  \edef\sdaau@temp{\expandafter\sdaau@getfive\meaning#1TTTTT$}%
  \ifx\sdaau@temp\sdaau@skipstring
    \typeout{\string\providelength: \string#1 already a length}%
  \else
    \@latex@error
      {\string#1 illegal in \string\providelength}
      {\string#1 is defined, but not with \string\newlength}%
  \fi
}
\def\sdaau@getfive#1#2#3#4#5#6${#1#2#3#4#5}
\edef\sdaau@skipstring{\string\skip}
\makeatother

\usepackage[capitalize]{cleveref}
\crefname{section}{Sec.}{Secs.}
\Crefname{section}{Section}{Sections}
\Crefname{table}{Table}{Tables}
\crefname{table}{Tab.}{Tabs.}

\def\xi{\mathbf{x}_i}

\graphicspath{{[CVPR 2023] Template/figures}, {examples}}

\makeatletter
 
\def\@fnsymbol#1{\ensuremath{\ifcase#1\or \dagger\or \ddagger\or
\mathsection\or \mathparagraph\or \|\or **\or \dagger\dagger
\or \ddagger\ddagger \else\@ctrerr\fi}}
\makeatother

\graphicspath{{figures/}}

\usepackage[pagenumbers]{cvpr} %
\makeatletter
\@namedef
{ver@everyshi.sty}{}
\makeatother
\usepackage{amsmath}
\usepackage[accsupp]{axessibility}

\def\cvprPaperID{3959} %
\def\confName{CVPR}
\def\confYear{2023}

\begin{document}

\title{
$\text{DC}^2$: Dual-Camera Defocus Control by Learning to Refocus 
}

\author{Hadi Alzayer\textsuperscript{\rm 1,2}~~~  Abdullah Abuolaim\textsuperscript{\rm 1}~~~ Leung Chun Chan\textsuperscript{\rm 1} \\
Yang Yang\textsuperscript{\rm 1}~~~ Ying Chen Lou\textsuperscript{\rm 1}~~~  Jia-Bin Huang\textsuperscript{\rm 2}~~~   Abhishek Kar\textsuperscript{\rm 1}\hfill \\ \\
\textsuperscript{\rm 1}Google \quad \quad
\textsuperscript{\rm 2}University of Maryland, College Park\\
}

\twocolumn[{
\renewcommand\twocolumn[1][]{#1}
\maketitle
\setlength{\fboxsep}{0pt}%
\setlength{\fboxrule}{1pt}%

\begin{center}
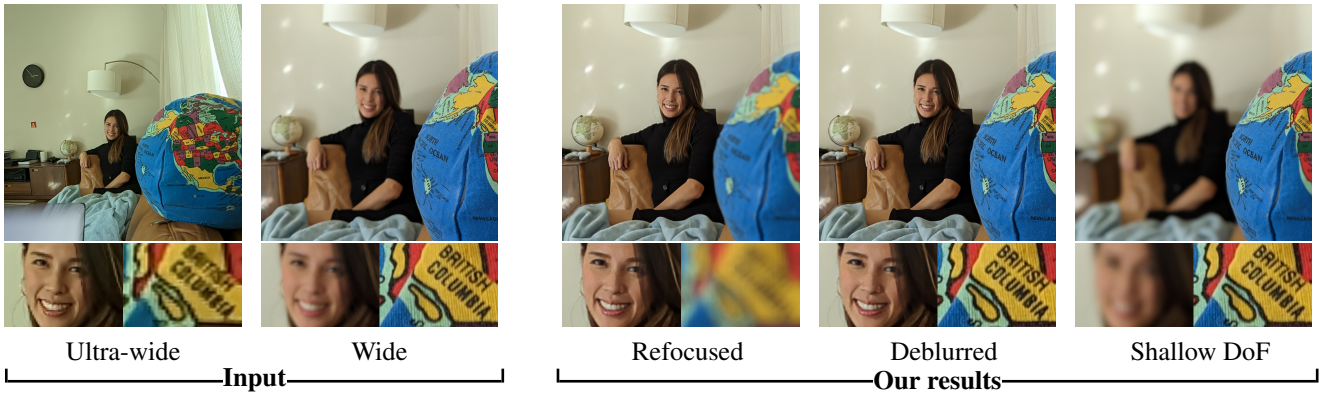

\mfigure{0.18}{teaser_person/sample_model_uw_cropped.png}
\mfigure{0.18}{teaser_person/sample_src_w_cropped.png} \hfill
\mfigure{0.18}{teaser_person/sample_model_output_refocus_cropped.png} 
\mfigure{0.18}{teaser_person/sample_model_output_deblurring_cropped.png}
\mfigure{0.18}{teaser_person/sample_model_output_shallow_cropped.png}

\mfigure{0.18}{teaser_person/stacked_uw.png} 
\mfigure{0.18}{teaser_person/stacked_w.png} \hfill
\mfigure{0.18}{teaser_person/stacked_refocused.png}
\mfigure{0.18}{teaser_person/stacked_deblurred.png}
\mfigure{0.18}{teaser_person/stacked_shallow.png}

\mpage{0.18}{Ultra-wide}
\mpage{0.18}{Wide} \hfill
\mpage{0.18}{Refocused}
\mpage{0.18}{Deblurred} 
\mpage{0.18}{Shallow DoF} \\
\vspace{-3mm}

$\underbracket[1pt][1.5mm]{\hspace{0.38\linewidth}}_%
    {\substack{\vspace{-4.0mm}\\\colorbox{white}
    {\textbf{Input}}}}$
\vspace{2mm} \hfill
$\underbracket[1pt][1.5mm]{\hspace{0.58\linewidth}}_%
    {\substack{\vspace{-4.0mm}\\\colorbox{white}
    {\textbf{Our results}}}}$

\captionof{figure}{
\textbf{Post-capture depth-of-field (DoF) control from dual camera.} 
(\emph{Left}) Using photos captured from dual cameras with a wide field of view (FoV) and ultra-wide FoV, our method enables various DoF manipulations. 
(\emph{Right}) We showcase our results of refocusing (changing the focal plane), deblurring (creating all-in-focus imagery), and synthesizing a shallower DoF (producing bokeh effects). 
}
\label{fig:teaser}
\end{center}
}]

\maketitle

\begin{abstract}
\vspace{-3mm}
Smartphone cameras today are increasingly approaching the versatility and quality of professional cameras through a combination of hardware and software advancements. However, fixed aperture remains a key limitation, preventing users from controlling the depth of field (DoF) of captured images. At the same time, many smartphones now have multiple cameras with \textit{different} fixed apertures - specifically, an ultra-wide camera with wider field of view and deeper DoF and a higher resolution primary camera with shallower DoF. In this work, we propose $\text{DC}^2$, a system for \textbf{defocus control} for synthetically varying camera aperture, focus distance and arbitrary defocus effects by fusing information from such a dual-camera system. Our key insight is to leverage real-world smartphone camera dataset by using image refocus as a proxy task for learning to control defocus. Quantitative and qualitative evaluations on real-world data demonstrate our system's efficacy where we outperform state-of-the-art on defocus deblurring, bokeh rendering, and image refocus. Finally, we demonstrate creative post-capture defocus control enabled by our method, including tilt-shift and content-based defocus effects.
\vspace{-3mm}
\end{abstract}

\section{Introduction}
\label{sec:intro}

Smartphone cameras are the most common modality for capturing photographs today~\cite{Dx0_article}. 
Recent advancements in computational photography such as burst photography~\cite{hdrplus_siggraph}, synthetic bokeh via portrait mode~\cite{googlePortraitMode}, super-resolution~\cite{burst_superres_siggraph}, and more have been highly effective at closing the gap between professional DSLR and smartphone photography. 
However, a key limitation for smartphone cameras today is depth-of-field (DoF) control, i.e., controlling parts of the scene that appear in (and out of) focus. 
This is primarily an artifact of their relatively simple optics and imaging systems (e.g., fixed aperture, smaller imaging sensors, etc.). 
To bridge the gap, modern smartphones tend to computationally process the images for further post-capture enhancements such as synthesizing shallow DoF (e.g., portrait mode~\cite{googlePortraitMode,Peng2022BokehMe}). 
However, this strategy alone does not allow for DoF \textit{extension} or post-capture refocus. 
In this work, we propose Dual-Camera Defocus Control ($DC^2$), a framework that can provide post-capture \textbf{defocus control} leveraging multi-camera systems prevalent in smartphones today. Figure~\ref{fig:teaser} shows example outputs from our framework for various post-capture DoF variations. In particular, our method is controllable and enables image refocus, DoF extension, and reduction.

Post-capture defocus control is a compound process that involves removing defocus blur (i.e., defocus deblurring) and then adding defocus blur selectively based on the scene depth. Defocus deblurring~\cite{karaali2018nonblind, shi2015justnoticable, ruan2022learning, Lee2021IFAN, Son_2021_ICCV, ruan2021TCI, ma2022tip, abuolaim2022improving, abuolaim2021learning, Xu_2021_CVPRW_EDPN, Pan_2021_CVPR, Xin_2021_ICCV_dual_pixel, abuolaim2020defocus}, itself, is challenging due to the nature of the defocus point spread function (PSF) formation which can be spatially varying in size and shape~\cite{levin2011understanding,tang2012utilizing}. 
The PSF's shape and size are not only depth dependent, but also vary based on aperture size, focal length, focus distance, optical aberration, and radial distortion. Synthesizing and adding defocus blur~\cite{hach2015blur, barron2015bilateral, googlePortraitMode, purohit2019bokeh, xiao2018deepfocus, xian2021salient, ignatov2020rendering, Nagasubramaniam2022, Peng2022BokehMe} is also difficult and requires an accurate depth map along with an all-in-focus image. 
Additionally, it requires realistic blur formation and blending around the object's boundaries. 
Most prior work has addressed defocus deblurring and synthesizing defocus blur as two isolated tasks. 
There has been less work on post-capture defocus control (e.g., image refocusing~\cite{refocusGAN, Ng2005LightFP, Jacobs2012FocalSC}). 
The image refocusing literature~\cite{Ng2005LightFP, Jacobs2012FocalSC} has focused on light-field data captured with specialized hardware. While the results in~\cite{wang2021DCSR, wang2017light} are the state-of-the-art, light-field data is not representative of smartphone and DSLR cameras by lacking realistic defocus blur and spatial resolution~\cite{boominathan2014improving}.

Most modern smartphones are now equipped with two or more rear cameras to assist with computational imaging. The primary camera -- often referred to as the wide camera or $\mathbf{W}$ -- has a higher resolution sensor, a higher focal length lens but a relatively shallower DoF. 
Alongside $\mathbf{W}$ is the ultra-wide ($\mathbf{UW}$) camera, often with a lower resolution sensor, lower focal length (wider field of view) and wider DoF. 
Our critical insight is to leverage this unique camera setup and cross-camera DoF variations to design a system for realistic post-capture defocus control. 
Differently from prior work, we tackle the problem of defocus control (deblurring \textit{and} adding blur) and propose using real-world data easily captured using a smartphone device to train our learning-based system. Our primary contributions in this work are as follows:
\begin{itemize}
    \item We propose a learning-based system for \textbf{defocus control} on dual-camera smartphones. This subsumes the tasks of defocus deblurring, depth-based blur rendering, image refocusing and enables arbitrary post-capture defocus control.
    \item In the absence of defocus control ground-truth, we enable training our system on real-world data captured from a smartphone device. To achieve that, we re-formulate the problem of defocus control as learning to refocus and define a novel training strategy to serve the purpose.
    \item We collect a dataset of diverse scenes with focus stack data at controlled lens positions the $\mathbf{W}$ camera and accompanying $\mathbf{UW}$ camera images for training our system. 
    Additionally, we compute all-in-focus images using the focus stacks to quantitatively evaluate image refocus, defocus deblurring and depth-based blurring tasks and demonstrate superior performance compared to state-of-the-art (SoTA) methods across all three tasks.
    \item Finally, we demonstrate creative defocus control effects enabled by our system, including tilt-shift and content-based defocus.
\end{itemize}

\begin{figure*}[t!]
    \centering
\mfigure{0.32}{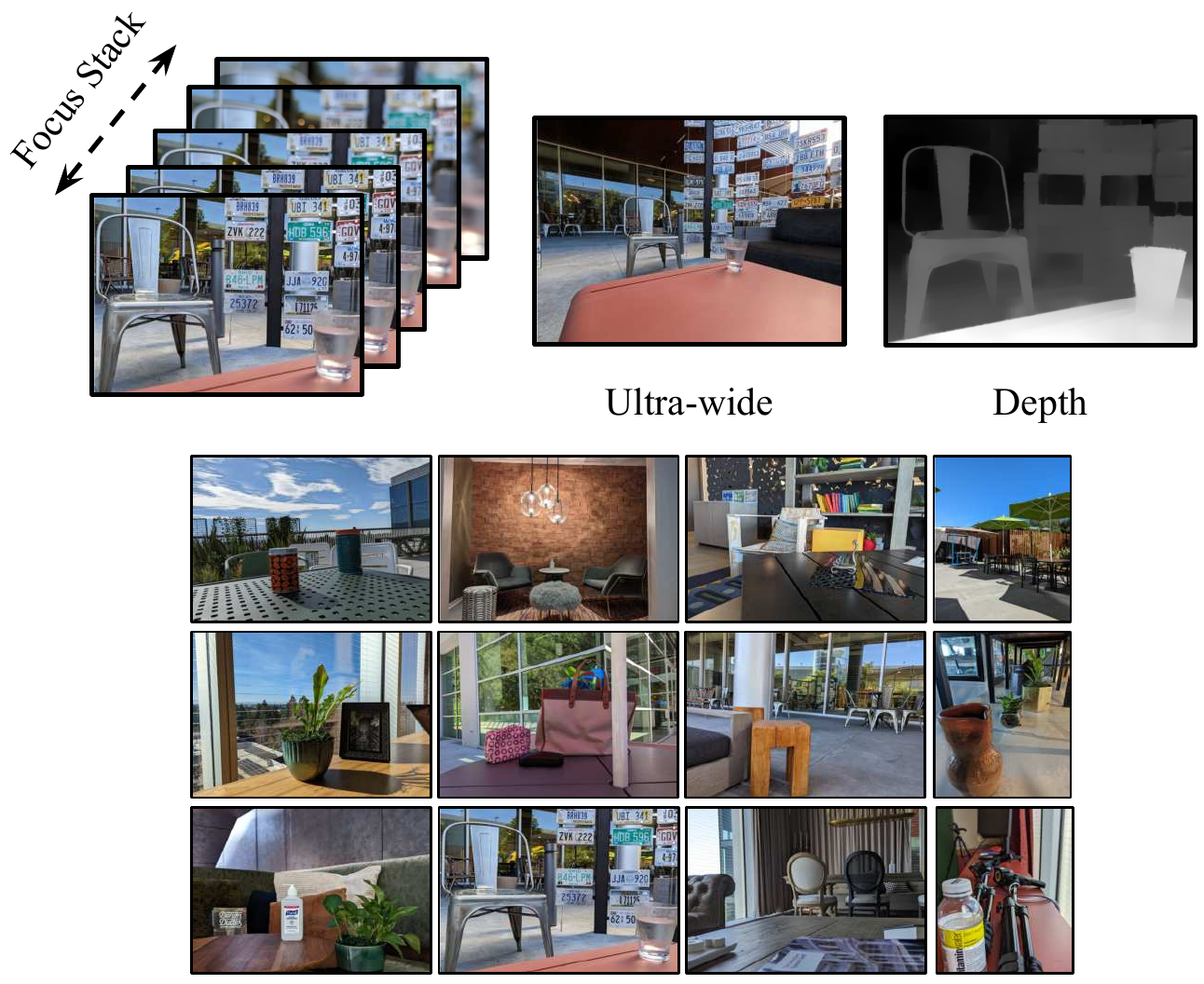} \hfill\vline\hfill 
\mfigure{0.28}{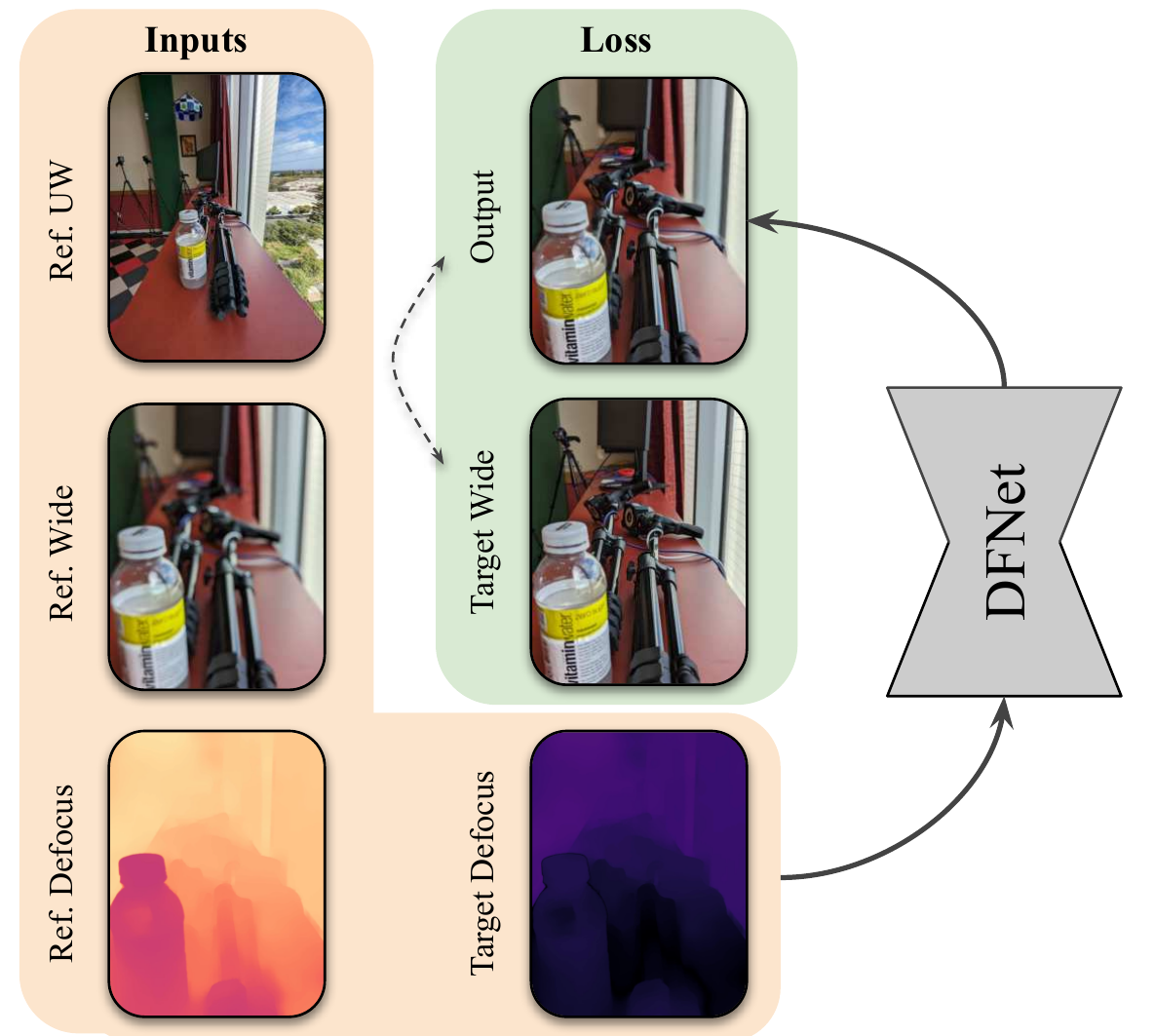} \hfill\vline\hfill
\mfigure{0.35}{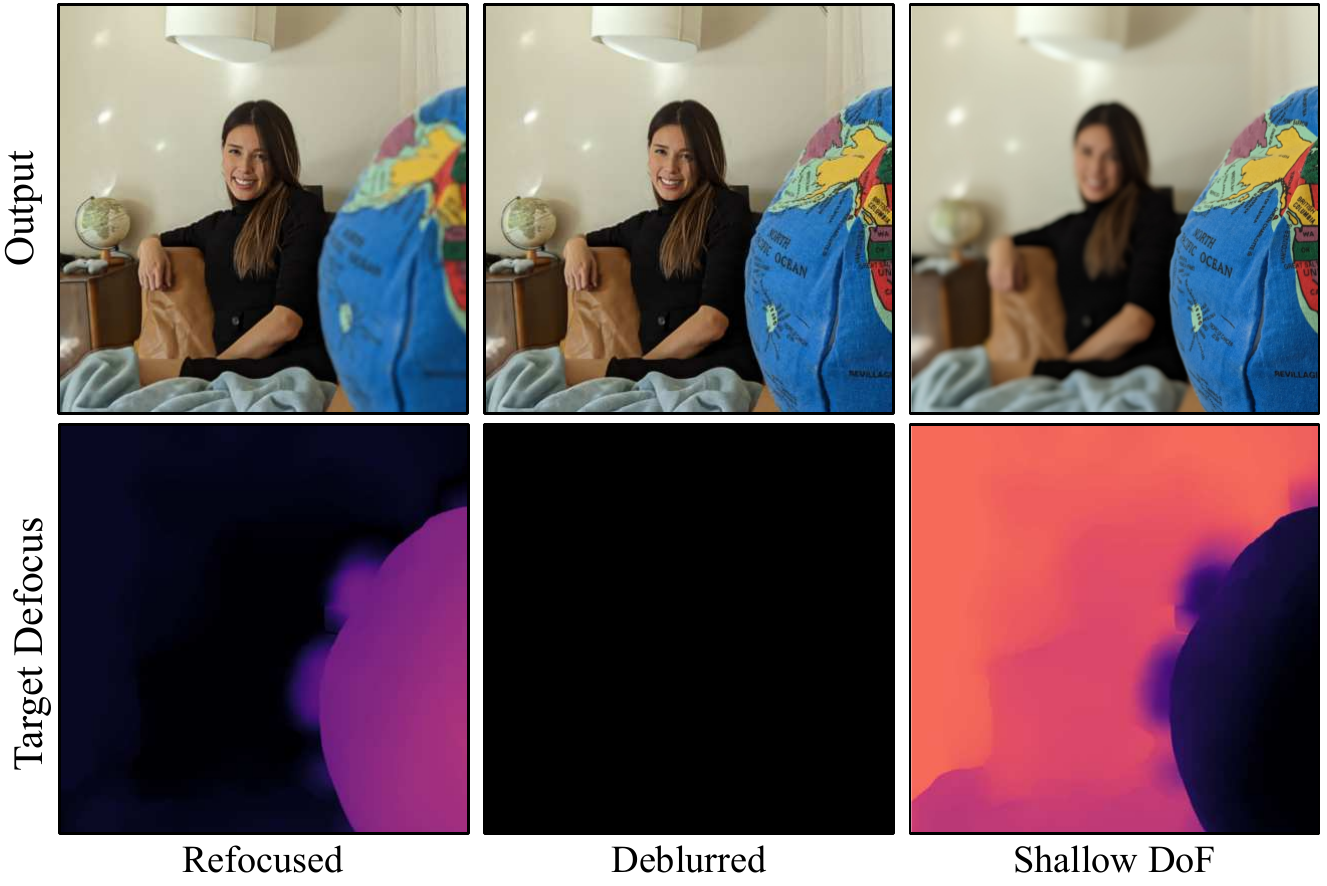} 
\mpage{0.32}{(a) Dual camera image refocus dataset.}
\mpage{0.28}{(b) Proxy task: Learn to refocus.} 
\mpage{0.35}{(c) Evaluate on arbitrary defocus maps.}
\caption{
\textbf{Image refocus as a proxy task.} 
Since we cannot gather a \textit{real} dataset for arbitrary focus manipulation, our idea is to train a model to perform \emph{image refocus} using a target defocus map as an input. 
At the test time, our trained model can perform arbitrary focus manipulation by feeding it an arbitrary target defocus map.
}

    \label{fig:refocus_idea}
\end{figure*}

\section{Related Work}
\label{sec:related}

\topic{Defocus Deblurring} 
Defocus blur leads to a loss of detail in the captured image.
To recover lost details, a line of work follows a two-stage approach: 
(1) estimate an explicit defocus map, 
(2) use a non-blind deconvolution guided by the defocus map \cite{ karaali2018nonblind, shi2015justnoticable}. 
With the current advances in learning-based techniques, recent work perform single image deblurring directly by training a neural network end-to-end to restore the deblurred image \cite{ruan2022learning, Lee2021IFAN, Son_2021_ICCV, ruan2021TCI, ma2022tip, abuolaim2022improving}. Due to the difficulty of the defocus deblurring task, other works try to utilize additional signals, such as the dual pixel (DP) data to improve deblurring performance \cite{abuolaim2021learning, Xu_2021_CVPRW_EDPN, Pan_2021_CVPR, Xin_2021_ICCV_dual_pixel, abuolaim2020defocus}. 
DP data is useful for deblurring as it provides the model with defocus disparity that can be used to inform deblurring. 
While the DP data provides valuable cues for the amount of defocus blur at each pixel, the DP views are extracted from a single camera. Therefore, the performance of the DP deblurring methods drops noticeably and suffer from unappealing visual artifacts for severely blurred regions. \\
In the same vein, we aim to exploit the \textbf{UW} image as a complementary signal already available in modern smartphones yet ignored for DoF control. By using the \textbf{UW} image with different DoF arrangements, we can deblur regions with severe defocus blur that existing methods cannot handle because of the fundamental information loss. Nevertheless, we are aware that using another camera adds other challenges like image misalignment, occlusion, and color mismatches which we address in Section \ref{sec:method}.

\topic{Bokeh Rendering} 
Photographers can rely on shallow DoF to highlight an object of interest and add an artistic effect to the photo. The blur kernel is spatially variant based on depth as well as the camera and optics. To avoid the need of estimating depth, one work magnifies the existing defocus in the image to make the blur more apparent without explicit depth estimate  \cite{bae2007defocus}. Since recent work in depth estimation improved significantly \cite{MegaDepthLi18, Srinivasan_2018_CVPR}, many shallow DoF rendering methods assume having depth \cite{Peng2022BokehMe} or estimate depth in the process \cite{googlePortraitMode, xian2021salient}. Using an input or estimated depth map, a shallow DoF can be synthesized using classical rendering methods \cite{hach2015blur, barron2015bilateral, googlePortraitMode, purohit2019bokeh}, using a neural network to add the synthetic blur \cite{deepLens, ignatov2020rendering, Nagasubramaniam2022} or a combination of classical and neural rendering \cite{Peng2022BokehMe}. With that said, shallow DoF synthesis methods typically assume an all-in-focus image or an input with a deep DoF.\\
Our proposed framework learns to blur as a byproduct of learning to refocus with the insight that the refocus task involves both deblurring and selective blurring. Unlike prior work that addressed either defocus deblurring or image bokeh rendering, we introduce a generic framework that facilitates post-capture full defocus control (e.g., image refocusing).

\topic{Image Refocus and DoF Control} 
At capture time, the camera focus can be adjusted automatically (i.e., autofocus~\cite{abuolaim2018revisiting,abuolaim2020online,herrmann2020learning}) or manually by moving the lens or adjusting the aperture. 
When the image is captured, it can still be post-processed to manipulate the focus. Nevertheless, post-capture image refocus is challenging as it requires both deblurring and blurring. Prior work uses specialized hardware to record a light field which allows post-capture focus control~\cite{Ng2005LightFP, Wang_2019}. 
However, light field cameras have low spatial resolution and are not representative of smartphone cameras. 
An alternative to requiring custom hardware is to capture a focus stack, and then merge the frames required to simulate the desired focus distance and DoF \cite{Jacobs2012FocalSC, paul2016multi, li2001combination, bhat2021multi}, but the long capture time restricts using focus stacks to static scenes. 
Research on single-image refocus is limited due to its difficulty, but the typical approach is to deblur to obtain an all-in-focus image followed by blurring. 
Previous work used classical deblurring and blurring \cite{classicRefocus} to obtain single image refocus, and the most notable recent single-image-based image refocus is RefocusGAN~\cite{refocusGAN}, which trains a two-stages GAN to perform refocusing. 
The limited research on software-based image refocus is likely due to the challenging task that involves both defocus deblurring and selective blurring. 
In our work, we provide a practical setup for post-capture image refocus without the restrictions of inaccessible hardware or the constraint of capturing a focus stack. We do so by leveraging the dual camera that is available in modern smartphones.

\topic{Image Fusion.} 
Combining information from images with complementary information captured using different cameras \cite{paul2016multi, multiviewfusiongoogle} or the same camera with different capture settings \cite{debevec2008recovering, hdrplus_siggraph} can enhance images in terms of sharpness \cite{Jacobs2012FocalSC, multiviewfusiongoogle, paul2016multi}, illuminant estimation~\cite{abdelhamed2021leveraging}, exposure~\cite{debevec2008recovering, hdrplus_siggraph, paul2016multi, bilcu2008high}, or other aspects \cite{multiviewfusiongoogle, wang2019stereoscopic, Godard_2018_ECCV, mildenhall2018kpn}. 
With the recent prevalence of dual-camera smartphones today, researchers have pursued works that target this setup. 
One line of work has used dual-camera for super-resolution to take advantage of the different resolutions the cameras have in still photos~\cite{wang2021DCSR, SelfDZSR, WU2022116825} as well as in videos~\cite{Lee2022RefVSR}. 
The dual-camera setup has also been used in multiple commercial smartphones, e.g., Google Pixel devices to deblur faces by capturing an ultra-wide image with faster shutter time and fusing with the wide photo~\cite{face_unblur}. 
To our knowledge, we are the first to investigate using the dual-camera setup for defocus control.

\def\D{\altmathcal{D}}
\def\I{\altmathcal{I}}
\def\O{\altmathcal{O}}
\def\res{\altmathcal{R}}

\def\b{\mathbfit{b}}
\def\c{\mathbfit{c}}
\def\d{\mathbfit{d}}
\def\o{\mathbfit{o}}
\def\p{\mathbfit{p}}
\def\t{\mathbfit{t}}
\def\x{\mathbfit{x}}
\def\z{\mathbfit{z}}

\def\K{\mathbfit{K}}
\def\R{\mathbfit{R}}

\def\ang{\phi}
\def\dehom{\mu}
\def\proj{\pi}
\def\sigmoid{S}
\def\vis{\nu}
\def\r{\mathbfit{r}}

\def\bp{(\p\!)} %
\def\bt{(t\!)} %
\def\bx{(\x\neg)} %

\def\ok{\o_{\neg k}}
\def\tk{\t_{\neg k}}
\def\wk{w_{\neg k}}
\def\xi{\x_{\neg i}}
\def\zk{\z_{\neg k}}
\def\Kk{\K_{\neg k}}
\def\Rk{\R_{\neg k}}

\def\ng{\hspace{-0.1mm}}
\def\neg{\hspace{-0.2mm}}
\def\pos{\hspace{0.2mm}}

\makeatletter
\newcommand*\MY@rightharpoonupfill@{%
    \arrowfill@\relbar\relbar\rightharpoonup
}
\newcommand*\overrightharpoon{%
    \mathpalette{\overarrow@\MY@rightharpoonupfill@}%
}
\makeatother

\newlength{\depthofsumsign}
\setlength{\depthofsumsign}{\depthof{$\sum$}}
\newcommand{\nsum}[1][1.4]{%
    \mathop{%
        \raisebox
            {-#1\depthofsumsign+1\depthofsumsign}
            {\scalebox
                {#1}
                {$\displaystyle\sum$}%
            }
    }
}

\begin{figure*}[ht!]
    \centering
   \centering
    \mfigure{0.55}{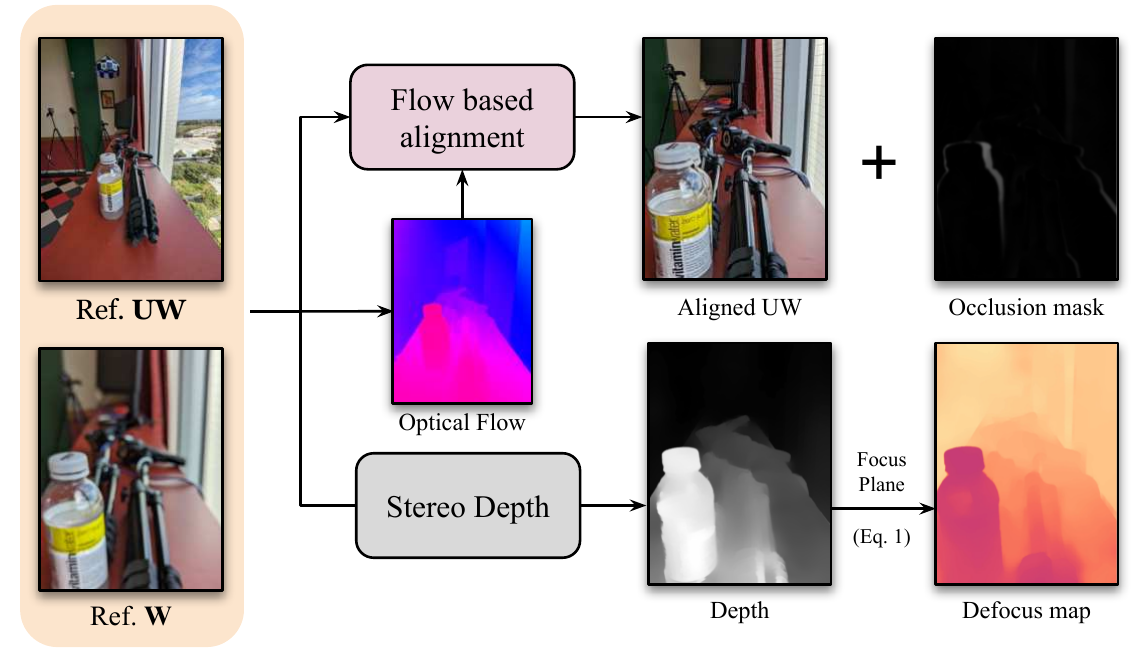} \hfill\vline\hfill
\mfigure{0.4}{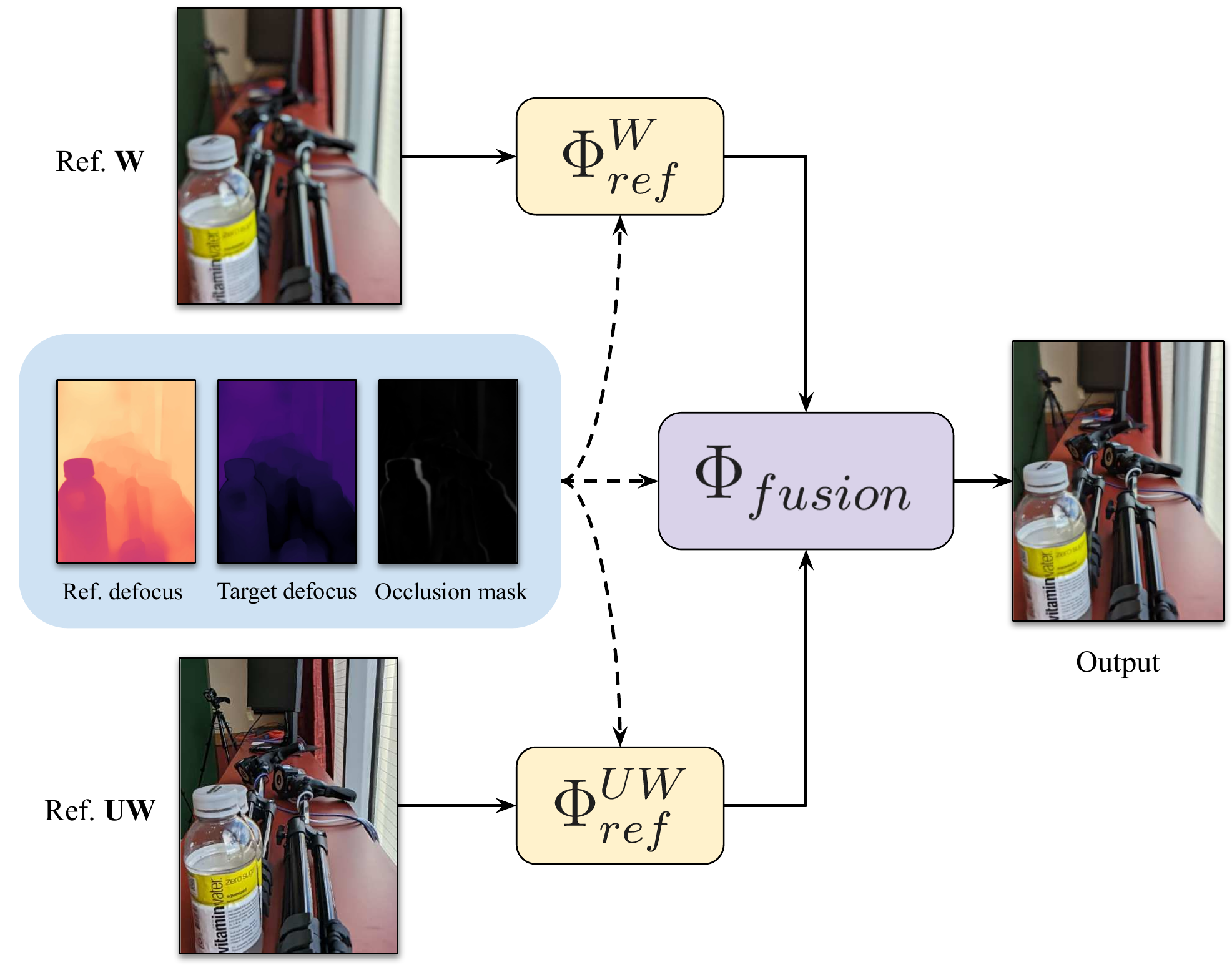}
    \mpage{0.55}{(a) Data processing} \hfill 
    \mpage{0.4}{(b) DFNet}
    \caption{
\textbf{Data processing and high-level architecture.} 
(\textit{Left}) To be able to use the reference inputs for our Detail Fusion Network, we need to align the inputs and a depth estimate to approximate the defocus map of the reference \textbf{W} and the target defocus map we would like to synthesize. 
We use flow-based alignment with PWCNet \cite{Sun2018PWC-Net} and use the stereo depth estimated using portrait mode \cite{googlePortraitMode}.
(\textit{Right}) Our Detail Fusion Network (DFNet) consists of refinement modules to refine the reference inputs combined with a fusion module that predicts blending masks to combine the two refined inputs.
}
    \label{fig:method}
\end{figure*}

\section{Learning to Refocus as a Proxy Task}
As mentioned, smartphone cameras tend to have fixed apertures limiting DoF control at capture time. 
In our work, we aim to unlock the ability to synthetically control the aperture - by transferring sharper details where present and synthesizing realistic blur. 
However, to train such a model, we run into a chicken and egg problem: we require a dataset of images captured with different apertures, which isn't possible with smartphones. 
An alternative solution could be to generate such a dataset synthetically, but modeling a realistic point spread function (PSF) for the blur kernel is non-trivial \cite{abuolaim2021learning}. 
Professional DSLRs provide yet another alternative~\cite{ignatov2017dslr} but often require paired captures smartphone / DLSR captures to reduce the domain gap. Ideally, we would like to use the same camera system for both training and evaluation. 
To resolve this, we observe that a somewhat parallel task is image refocus. When we change the focus distance, the defocus radius is adjusted in different parts of the image, involving a combination of pixels getting deblurred and blurred. 
This suggests that image refocus is at least as hard as scaling the DoF. 
Motivated by this observation, we make the hypothesis that by training a model on image refocus as a \textit{proxy task}, we can use the same model to control the DoF at test time as we show in Figure \ref{fig:refocus_idea}. 
The key idea is to provide the model with reference and target defocus maps (~\secref{datacollection}) as input, and at test time control the model behavior by manipulating this target defocus map.

\section{Method}
To train a model on our proxy task, we need to collect a dataset of focus stacks for the wide camera and a paired ultra-wide frame which can be used as a guide due to its deeper DoF. In Figure \ref{fig:method} we show the high-level structure of our method dubbed $\text{DC}^2$. The primary module that we train is the Detail Fusion Network (DFNet), which requires a reference wide frame, (aligned) reference ultra-wide frame, and estimated defocus maps. 
In Section \ref{sec:datacollection}, we describe how we collect the focus stack data and process it to obtain the inputs needed for DFNet. We then describe the architecture details of DFNet in Section \ref{sec:model}, which is motivated by the dual-camera input setup.

\subsection{Data Processing}
\label{sec:datacollection}
Using the Google Pixel 6 Pro as our camera platform, we captured a dataset of 100 focus stacks of diverse scenes, including indoor and outdoor scenarios. 
For each scene, we sweep the focus plane for the wide camera and capture a complete focus stack. We simultaneously capture a frame from the ultra-wide camera, which has a smaller aperture, deeper DoF, and fixed focus. 
For each frame, we use optical-flow-based warping using PWCNet~\cite{Sun2018PWC-Net} and following prior work~\cite{face_unblur} to align the ultra-wide frame with the wide frame. Since the alignment is imperfect (e.g., in textureless regions and occluded boundaries), we estimate an occlusion mask that can be used to express potentially misaligned regions for the model.
To estimate defocus maps, we require the metric depth. We use the depth map embedded in the Pixel camera's portrait mode output which can estimate metric depth using dual camera stereo algorithms~\cite{du2net-zhang} with a known camera baseline. To compute the defocus map associated with each frame, we use the following formula for the radius of the circle of confusion $c$
\begin{align}
    c = A \frac{|S_2 - S_1|}{S_2} \frac{f}{S_1 - f}
\end{align}
where $A$ is the camera aperture, $S_1$ is the focus distance, $S_2$ is the pixel depth, and $f$ is the focal length. In Figure \ref{fig:refocus_idea}a, we show a visualization of a focus stack, associated \textbf{UW}, stereo depth, and a collection of sample scenes.

\subsection{Model Architecture}
\label{sec:model}

Our method performs detail fusion on two primary inputs: the reference wide (\textbf{W}) and ultra-wide (\textbf{UW}) images. Since we train the model to refocus, \textbf{W} is expected to be treated as a base image, while \textbf{UW} is a guide for missing high-frequency details.

Based on this intuition, we propose \textbf{Detail Fusion Network (DFNet)} that has two refinement paths: \textbf{W} refinement path ($\Phi_{ref}^W$), \textbf{UW} refinement path ($\Phi_{ref}^{UW}$), and a fusion module ($\Phi_{fusion}$) that predicts blending masks for the refined \textbf{W} and refined \textbf{UW}. Note that the \textbf{W} refinement path never gets to see the \textbf{UW} frame and vice versa. We use a network architecture based on Dynamic Residual Blocks Network (DRBNet)~\cite{ruan2022learning} for our refinement modules with multi-scale refinements. For the fusion module, we use a sequence of atrous convolutions \cite{chen2017rethinking} for an increased receptive field and predict a blending mask for each scale. To preserve high-frequency details in the blending mask, we add upsampling layer and residual connections when predicting the blending mask of the larger scale. During training, we blend the outputs of $\Phi_{ref}^W$ and $\Phi_{ref}^{UW}$ and compute the loss for all scales for improved performance. In Figure \ref{fig:method} we show a high-level diagram of our architecture and how each component interacts with the others. 
By visualizing the intermediate outputs between our different modules, we observe that the network indeed attempts to maintain the low-frequency signal from\textbf{W} while utilizing high-frequency signals from \textbf{UW}. 
Please refer to the supplementary material for a detailed model architecture and a deeper analysis of model behavior and visualizations.

\subsection{Training Details}
\label{sec:training}
We train our model by randomly sampling slices from the focus stack in our training scenes. For each element in the batch, we randomly sample a training scene, and sample two frames to use as reference and target images, respectively. 
While we can approximate depth from all pairs, severely blurry frames can have unreliable depth. To address that, we use the stereo pair with the greatest number of matched features to use for the scene depth to compute the defocus maps. We train on randomly cropped 256x256 patches, using a batch size of 8, and a learning rate of $10^{-4}$ for 200k iterations, and then reduce the learning rate to $10^{-5}$ for another 200k iterations using Adam \cite{adam_optim}. Our reconstruction loss is a combination of ${L}_1$ loss on pixels and gradient magnitudes, SSIM loss \cite{ssim}, and perceptual loss \cite{zhang2018perceptual}. For a target wide frame $\textbf{W}_{tgt}$ and a model output $y$, the loss is

\begin{equation}
\begin{aligned}
    L_{total} &= {L}_1(\textbf{W}_{tgt}, y) + {L}_1(\nabla \textbf{W}_{tgt}, \nabla y) \\
    &L_{SSIM}(\textbf{W}_{tgt},y)) + L_{VGG}(\textbf{W}_{tgt},y)
\end{aligned}
\end{equation}
\label{sec:method}

\section{Experimental Results}
\label{sec:result}

We train our method to perform defocus control through training on the \textit{proxy task} of image refocus. As a result, our model can perform a variety of related defocus control tasks. Specifically, we evaluate our method on defocus deblurring, synthesizing shallow DoF, and image refocus.

\noindent\textbf{Evaluation metrics.} We use the standard signal processing metrics, i.e., the peak signal-to-noise ratio (PSNR) and the structural similarity index measure (SSIM). We also report the learned perceptual image patch similarity (LPIPS)~\cite{zhang2018unreasonable}.

\begin{table}[t]
\centering
\caption{
\textbf{Defocus deblurring evaluation.} Performance on generating all-in-focus images from a single slice in the focus stack. 
The best results are in bold numbers.
}
\begin{tabular}{lccc}
\toprule

\textbf{Method} & \textbf{PSNR $\uparrow$} & \textbf{SSIM $\uparrow$} & \textbf{LPIPS $\downarrow$} \\
\midrule
MDP~\cite{abuolaim2022improving} & 23.50 & 0.674 & 0.394 \\
IFAN~\cite{Lee2021IFAN} & 23.48 & 0.679 &  0.371\\
DRBNet~\cite{ruan2022learning} & 24.27 & 0.681 & 0.377\\
Ours & \textbf{24.79} & \textbf{0.704} & \textbf{0.351}\\
\bottomrule
\end{tabular}
\vspace{-0.05in}
\label{tab:deblurring}
\end{table}

\begin{figure*}[t!]
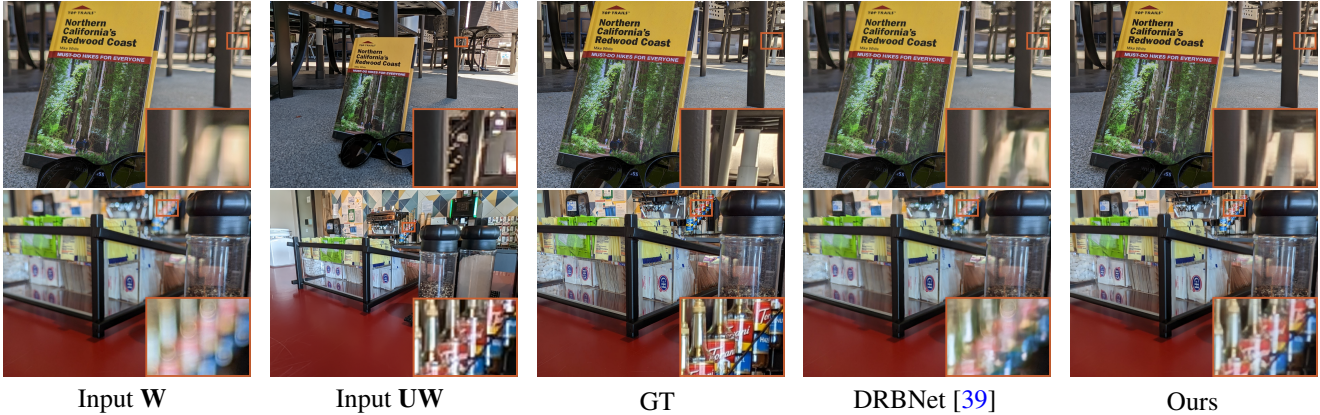

    \centering
        \mfigure{0.188}{deblurring_figures/book_src_w.jpg} \hfill
        \mfigure{0.188}{uw_images_resized/match_fov_uw_final_book_resized_modified.png}  \hfill
    \mfigure{0.188}{deblurring_figures/book_GT.jpg} \hfill 
    \mfigure{0.188}{deblurring_figures/book_drbnet.jpg} \hfill
    \mfigure{0.188}{deblurring_figures/book_ours.jpg}
    \mfigure{0.188}{deblurring_figures/kitchen_deblurring_src_w.jpg} \hfill
    \mfigure{0.188}{uw_images_resized/match_fov_uw_final_kitchen_resized_modified.png} \hfill
    \mfigure{0.188}{deblurring_figures/kitchen_deblurring_GT.jpg} \hfill
    \mfigure{0.188}{deblurring_figures/kitchen_deblurring_drbnet.jpg} \hfill
    \mfigure{0.188}{deblurring_figures/kitchen_deblurring_ours.jpg}
    \mpage{0.18}{Input \textbf{W}} \hfill 
    \mpage{0.18}{Input \textbf{UW}} \hfill 
    \mpage{0.18}{GT} \hfill 
    \mpage{0.18}{DRBNet \cite{ruan2022learning}} \hfill 
    \mpage{0.18}{Ours} 
    \caption{
\textbf{Defocus deblurring.} 
We showcase the results of our method against SoTA single image defcous deblurring DRBNet~\cite{ruan2022learning}. 
Note that our method restores severely blurred regions in the background that single-image based methods often struggle with.
}
    \label{fig:deblurring}
\end{figure*}

\subsection{Defocus Deblurring}
\noindent\textbf{Task.} 
The goal of defocus deblurring is to remove the defocus blur in the image. 
For our method to perform defocus deblurring, we simply set the target defocus map to all zeros. 
To obtain an all-in-focus image as a ground truth, we perform focus stacking using our focus stacks through commercial software provided by HeliconFocus. 
Then the evaluation task is deblurring individual slices from the focus stack to generate an all-in-focus image. 
Due to the focus magnification between the focus stack slices, we align the field-of-view (FoV) with the all-in-focus image through a combination of FoV matching and brute-force search for the best scaling and translation parameters that minimize the error. 
We use the same alignment method when evaluating all the methods to ensure fairness.

\noindent\textbf{Methods.} We compare our method with the following single-image defocus deblurring methods: Dynamic Residual Blocks Network (DRBNet)~\cite{ruan2022learning}, Multi-task DP (MDP) network~\cite{abuolaim2022improving}, and Iterative Filter Adaptive Network (IFAN) \cite{Lee2021IFAN}. Note that these methods do not take the ultra-wide image as input, and the main purpose of the comparison is to highlight the value of leveraging an available dual-camera setup. Our dataset does not contain DP data and thus we are not able to benchmark the DP defocus deblurring methods~\cite{abuolaim2021learning, Xu_2021_CVPRW_EDPN, Pan_2021_CVPR, Xin_2021_ICCV_dual_pixel, abuolaim2020defocus}. As for the evaluation on other defocus deblurring datasets (e.g.,~\cite{abuolaim2020defocus}), our method requires dual-camera input not available in current datasets.

\noindent\textbf{Evaluation.} 
In Table \ref{tab:deblurring}, we compare the performance of our method against other defocus deblurring methods. Our method achieves the best results on all metrics with dual camera inputs. Note that our method has never seen all-in-focus outputs / zero target defocus maps during training and learns to deblur via the proxy task. Figure \ref{fig:deblurring} shows two deblurring results of our method against DRBNet~\cite{ruan2022learning}. 
As shown in the zoomed-in insets, our method is able to restore severely blurred regions better compared to DRBNet. 
In general, single-image defocus deblurring methods suffer from artifacts and tend to hallucinate when restoring severely blurred regions. Therefore, an additional signal such as the \textbf{UW} is very useful when the details are completely lost in the input image. While the main task of our proposed method is not only defocus deblurring, it achieves the SoTA deblurring results quantitatively and qualitatively. 
These results also demonstrate how generic and flexible our proposed defocus control framework is.

\begin{figure*}[t!]
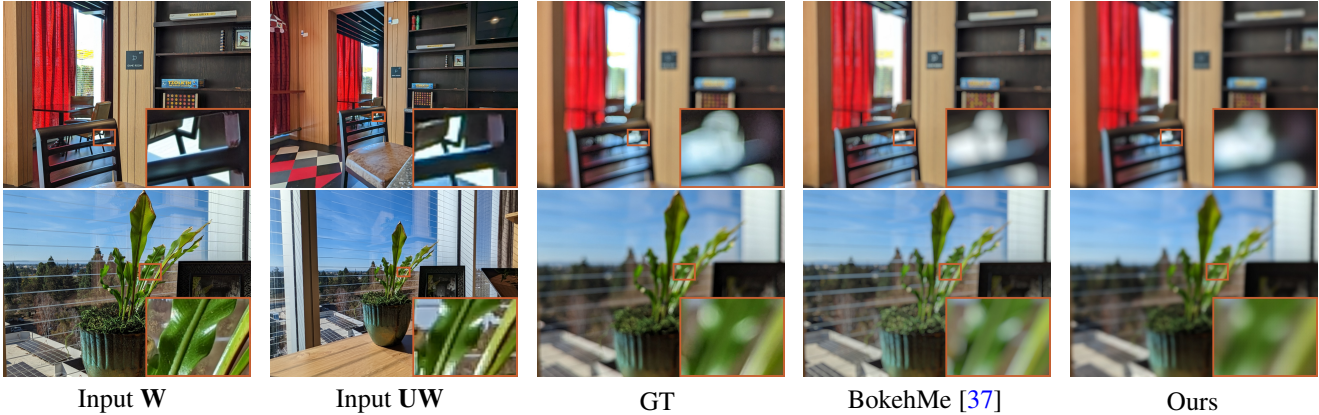

    \centering
\mfigure{0.188}{blurring_figures/gameroom_blur_GT.jpg} \hfill
\mfigure{0.188}{uw_images_resized/match_fov_uw_final_gameroom_resized_modified.png}  \hfill
\mfigure{0.188}{blurring_figures/gameroom_blur_src.jpg} \hfill
\mfigure{0.188}{blurring_figures/gameroom_blur_bokehme.jpg} \hfill
\mfigure{0.188}{blurring_figures/gameroom_blur_ours.jpg}
\mfigure{0.188}{blurring_figures/plant_window_image.jpg} \hfill
\mfigure{0.188}{uw_images_resized/match_fov_uw_final_glassplant_resized_modified.png}  \hfill
\mfigure{0.188}{blurring_figures/plant_window_src.jpg} \hfill
\mfigure{0.188}{blurring_figures/plant_window_bokehme.jpg} \hfill
\mfigure{0.188}{blurring_figures/plant_window_ours.jpg}
\mpage{0.18}{Input \textbf{W}} \hfill 
\mpage{0.18}{Input \textbf{UW}} \hfill 
\mpage{0.18}{GT} \hfill 
\mpage{0.18}{BokehMe \cite{Peng2022BokehMe}} \hfill 
\mpage{0.18}{Ours}

\caption{
\textbf{Blurring results.} 
Our method can synthesize shallow DoF from an all-in-focus image with a performance competitive with SoTA in bokeh rendering~\cite{Peng2022BokehMe}.
}
    \label{fig:blurring_piano}
\end{figure*}

\begin{table}[t]
\centering
\caption{
\textbf{Bokeh blurring evaluation.} performance on simulating different slices of the focus stack from the all-in-focus image.
}
\begin{tabular}{lccc}
\toprule
\textbf{Method} & \textbf{PSNR $\uparrow$} & \textbf{SSIM $\uparrow$} & \textbf{LPIPS $\downarrow$} \\
\midrule

BokehMe~\cite{Peng2022BokehMe} & 26.65 & 0.870 & 0.241\\
Neural Rend.~\cite{Peng2022BokehMe} & 27.87 & 0.874 & 0.246 \\
Classic Rend.~\cite{Peng2022BokehMe} & 26.66 & 0.870 & 0.241\\
Ours & \textbf{29.78} & \textbf{0.898} &\textbf{ 0.172}\\

\bottomrule
\end{tabular}

\vspace{-0.05in}
\label{tab:blurring}
\end{table}

\subsection{Shallow DoF Rendering}
\noindent \textbf{Task.} We also evaluate our method on rendering shallow DoF images. The input to the method is an all-in-focus image, an approximate target defocus map, and the desired output is the image with a synthetic shallow DoF guided by the defocus map. We use the all-in-focus image generated from the focus stack as input and try to reconstruct the various slices in the focus stack using each slice's defocus map as a target.

\noindent \textbf{Methods.}
We compare against BokehMe~\cite{Peng2022BokehMe}, a recent state-of-the-art in shallow DoF synthesis that relies on blending the outputs of classic blur synthesis with neural rendering methods. We also evaluate the classical scattering-based blur and the neural renderer within BokehMe in isolation.

\noindent \textbf{Evaluation.}
In Table \ref{tab:blurring}, we show that our method is competitive with SoTA shallow DoF rendering methods. Note that for DoF reduction, $\mathbf{UW}$ does not provide a useful signal since the task primarily involves signal removal from $\mathbf{W}$, but the model learns to perform this task as a byproduct of training on image refocus. In Figure \ref{fig:blurring_piano} we show visual results where our model synthesizes realistic blur.

\begin{figure*}[t!]
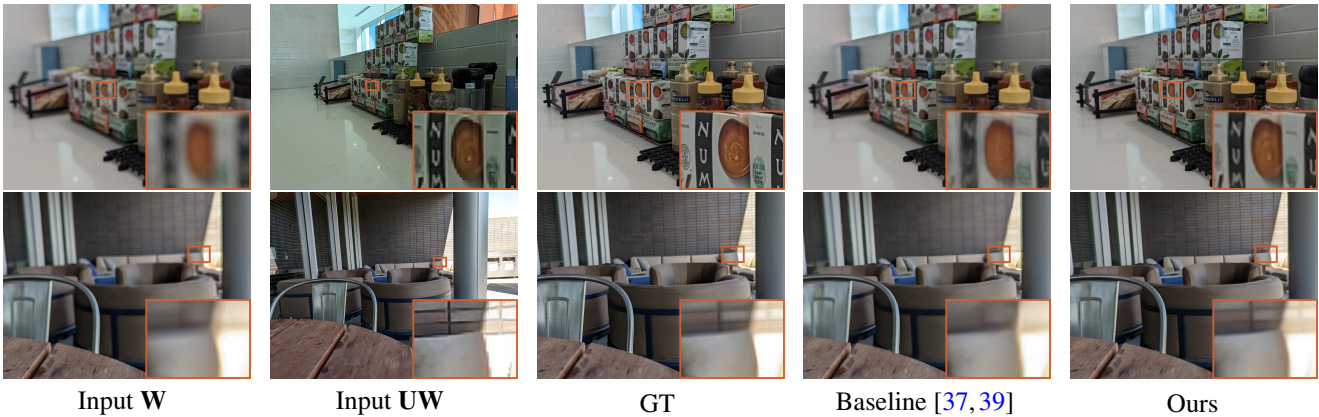

    \centering
\mfigure{0.188}{refocus_figures/tea_refocus_src.jpg} \hfill
\mfigure{0.188}{uw_images_resized/match_fov_uw_final_micro_resized_modified.png} \hfill
\mfigure{0.188}{refocus_figures/tea_refocus_target.jpg} \hfill
\mfigure{0.188}{refocus_figures/tea_refocus_baseline.jpg} \hfill
\mfigure{0.188}{refocus_figures/tea_refocus_ours.jpg} 
\mfigure{0.188}{refocus_figures/cafeteria_src.jpg} \hfill
\mfigure{0.188}{uw_images_resized/match_fov_uw_final_cafeteria_resized_modified.png} \hfill
\mfigure{0.188}{refocus_figures/cafeteria_target.jpg} \hfill
\mfigure{0.188}{refocus_figures/cafeteria_baseline.jpg} \hfill
\mfigure{0.188}{refocus_figures/cafeteria_ours.jpg}
\mpage{0.18}{Input \textbf{W}} \hfill 
\mpage{0.18}{Input \textbf{UW}} \hfill 
\mpage{0.18}{GT} \hfill 
\mpage{0.18}{Baseline \cite{ruan2022learning,Peng2022BokehMe}} \hfill 
\mpage{0.18}{Ours} 

\caption{\textbf{Refocus results.} 
We shift the focus plane and demonstrate that we can match the desired refocused image and the target blur without completely deblurring the input. 
Our method outperforms the baseline that refocuses by deblurring and reblurring the image.}
    \label{fig:refocus}
\end{figure*}

\begin{table}[t]
\centering
\caption{
\textbf{Image refocus evaluation.} Performance on re-synthesizing focus planes given an input with different focus plane from the same scene.}
\begin{tabular}{lccc}
\toprule
\textbf{Method} & \textbf{PSNR $\uparrow$} & \textbf{SSIM $\uparrow$} & \textbf{LPIPS $\downarrow$} \\
\midrule
UW + Blur~\cite{Peng2022BokehMe} & 21.89 & 0.803 & 0.364 \\
Deblur~\cite{ruan2022learning}+Reblur~\cite{Peng2022BokehMe} & 26.40 & 0.833 & 0.312 \\
Ours &  \textbf{28.58} &  \textbf{0.860} & \textbf{0.217}\\
\bottomrule
\end{tabular}
\vspace{-0.05in}
\label{tab:refocus}
\end{table}

\subsection{Image Refocus}
\noindent \textbf{Task.}
Image refocus involves shifting the focus plane and as a result, the near and far focus depths. To evaluate on image refocus, we randomly sample two frames from a focus stack, a reference frame, and a target frame, and evaluate the model performance in reproducing the target frame.

\noindent \textbf{Methods.}
There is limited work on single-image refocus, the most notable work being RefocusGAN~\cite{refocusGAN}. The idea behind RefocusGAN is to use generative models to deblur the image followed by blurring it. This approach is likely because of the difficulty of realistically switching between different defocus amounts directly~\cite{bae2007defocus}. However, we are not able to compare with RefocusGAN as the code and trained models are not available. As an alternative for comparison, we adopt SoTA in defocus deblurring (DRBNet~\cite{ruan2022learning}) and SoTA in blurring (BokehMe~\cite{Peng2022BokehMe}) for image refocus. We also compare against blurring the aligned \textbf{UW} directly since it could approximate an all-in-focus image.

\noindent \textbf{Evaluation.}
In Table \ref{tab:refocus} we show that our method outperforms the baseline in image refocus. Note that since we train our method to switch between the reference defocus to the target defocus, the model can implicitly learn to switch between different PSF scales from the data. We show visual results in Figure \ref{fig:refocus}. Note that when the target image contains blurry regions like shown on the wall, our method deblurs the input just enough to match the target defocus.

\begin{table}[t]
\centering
\caption{
\textbf{Ablations on Image Input}. Comparison on different input types. Although performance increases by removing the occlusion mask, qualitative performance drops (see Figure \ref{fig:ablation}).
}
\begin{tabular}{lccc}
\toprule
\textbf{Method} & \textbf{PSNR $\uparrow$} & \textbf{SSIM $\uparrow$} & \textbf{LPIPS $\downarrow$} \\
\midrule
W only & 28.44 & 0.855 & 0.260\\
UW only & 22.66 & 0.822 & 0.307\\
No occlusion & \textbf{28.81} & \textbf{0.864} & 0.219 \\
Full input &  28.58 &  0.860 & \textbf{0.217} \\
\bottomrule
\end{tabular}

\vspace{-0.05in}
\label{tab:template}
\end{table}

\begin{figure}[t!]
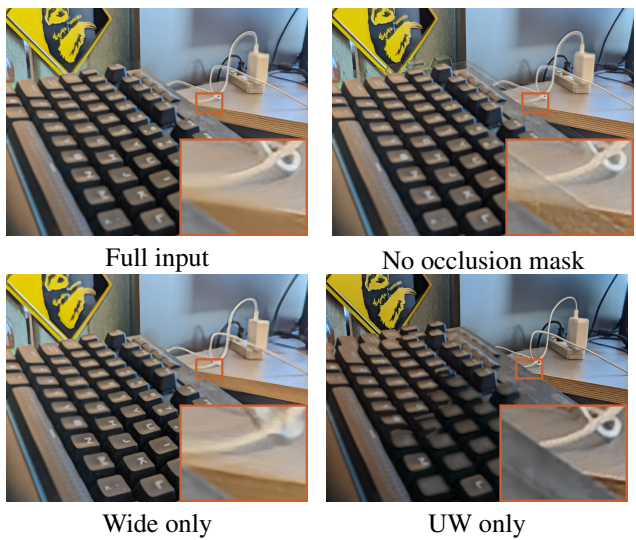

    \centering
    \mfigure{0.48}{ablation_figures/keyboard_full_modified.png} \hfill
    \mfigure{0.48}{ablation_figures/keyboard_no_occlusion_modified.png}
        \mpage{0.48}{Full input} \hfill 
    \mpage{0.48}{No occlusion mask} \hfill
        \mfigure{0.48}{ablation_figures/keyboard_wide_only_modified.png} \hfill
    \mfigure{0.48}{ablation_figures/keyboard_uw_only_modified.png}

    \mpage{0.48}{Wide only} \hfill 
    \mpage{0.48}{UW only} 
    \caption{
    \textbf{Ablation.} Using the occlusion mask helps the model avoid transferring warping artifacts to the final image, while using \textbf{W} only hinders deblurring performance, and \textbf{UW} only suffers from warping artifacts and lower resolution.}
    \label{fig:ablation}
\end{figure}

\subsection{Ablation Study}
\label{sec:ablation}
The key idea of our work is using the ultra-wide camera as a guide to performing DoF control. To evaluate the effects of using \textbf{UW}, we train a model using only W (only keeping the Wide refinement module) and similarly training a \textbf{UW} only model. We compare their performance on image refocus in Table \ref{tab:template}. Note that while the wide input is sufficient when the target involves only blurring or minimal deblurring, it is an ill-posed setup when it requires considerable deblurring. On the other hand, the warped \textbf{UW} lower quality severely limits the performance when relying on it completely. We visualize an example in In Figure \ref{fig:ablation}. Note that when using \textbf{W} only, deblurring performance is limited. Also we note that when removing the occlusion mask, while signal-processing metrics could see slight improvements, qualitative performance drops as we can observe ghosting artifacts around occluded boundaries.

\begin{figure}[t!]
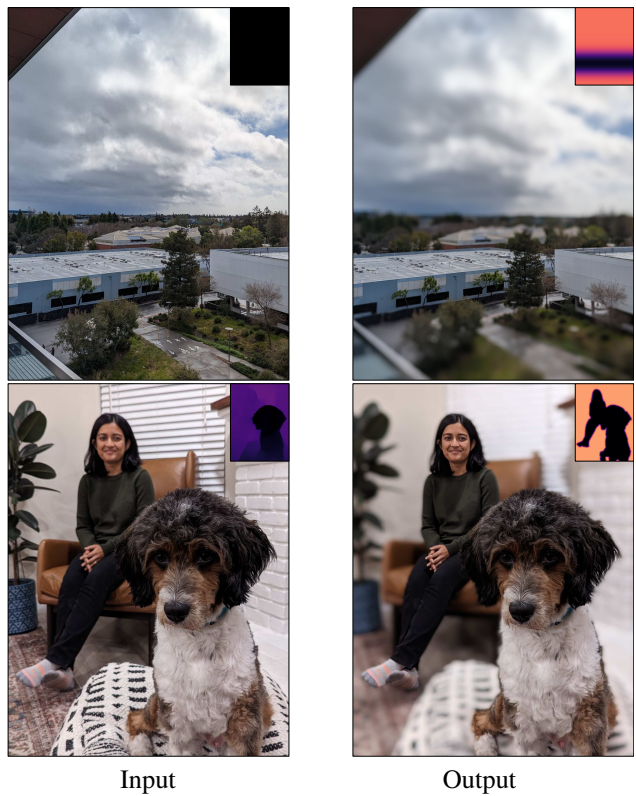

    \centering
    \mfigure{0.45}{application_figures/tiltshift_inp_large.pdf} \hfill
    \mfigure{0.45}{application_figures/tiltshift_out_large.pdf}
    \mfigure{0.45}{application_figures/seg_mask_input.pdf} \hfill
    \mfigure{0.45}{application_figures/seg_mask_out_2.pdf} \hfill
    \mpage{0.45}{Input} \hfill 
    \mpage{0.45}{Output} \hfill 
    \caption{
    \textbf{Creative applications}. (\textit{Top}) we apply tilt shift effect that makes large objects appear as miniatures. (\textit{Bottom}) we use segmentation mask to deblur objects of interest and blur the background.}
    \label{fig:application}
\end{figure}
\noindent \textbf{Applications.}
Our method allows for arbitrary target defocus maps as an input. In Figure \ref{fig:application} we demonstrate a \textit{tilt-shift} effect, where a large scene appears smaller because of the blur, as well as using a segmentation mask to deblur objects of interest (the person) while blurring the remaining objects.

\begin{figure}[t!]
    \centering
    \mfigure{0.45}{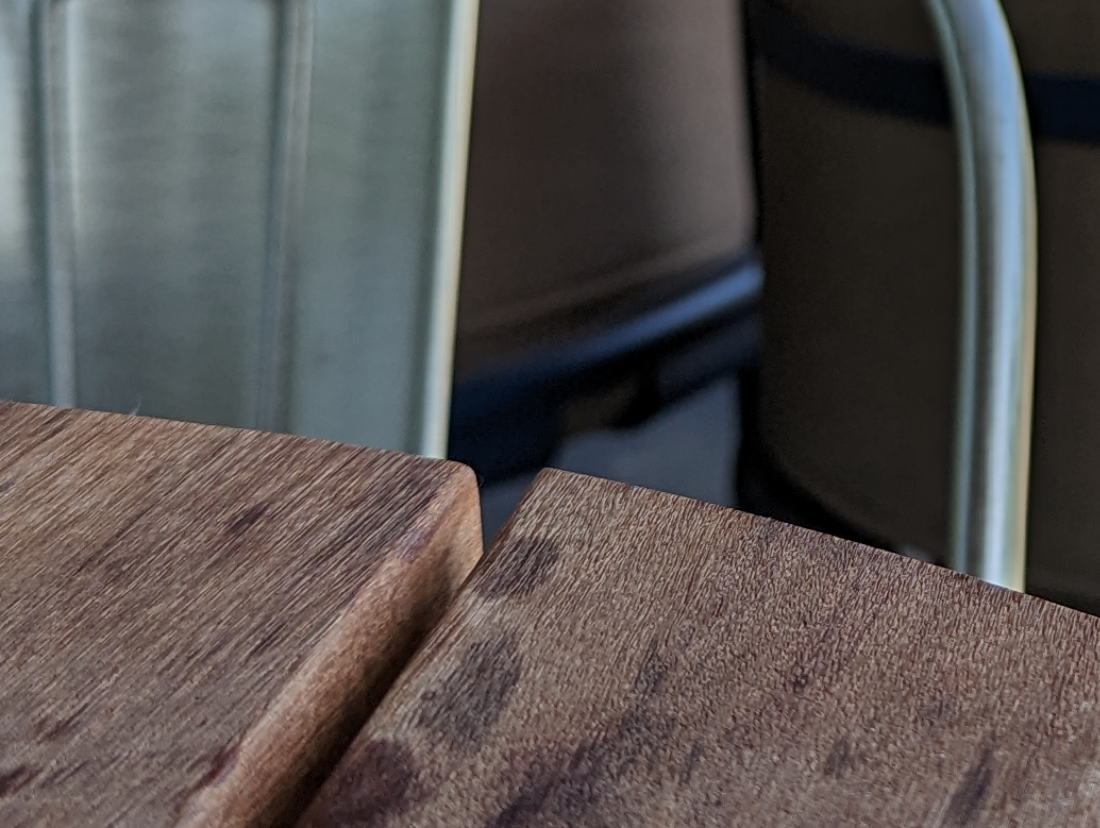} \hfill
    \mfigure{0.45}{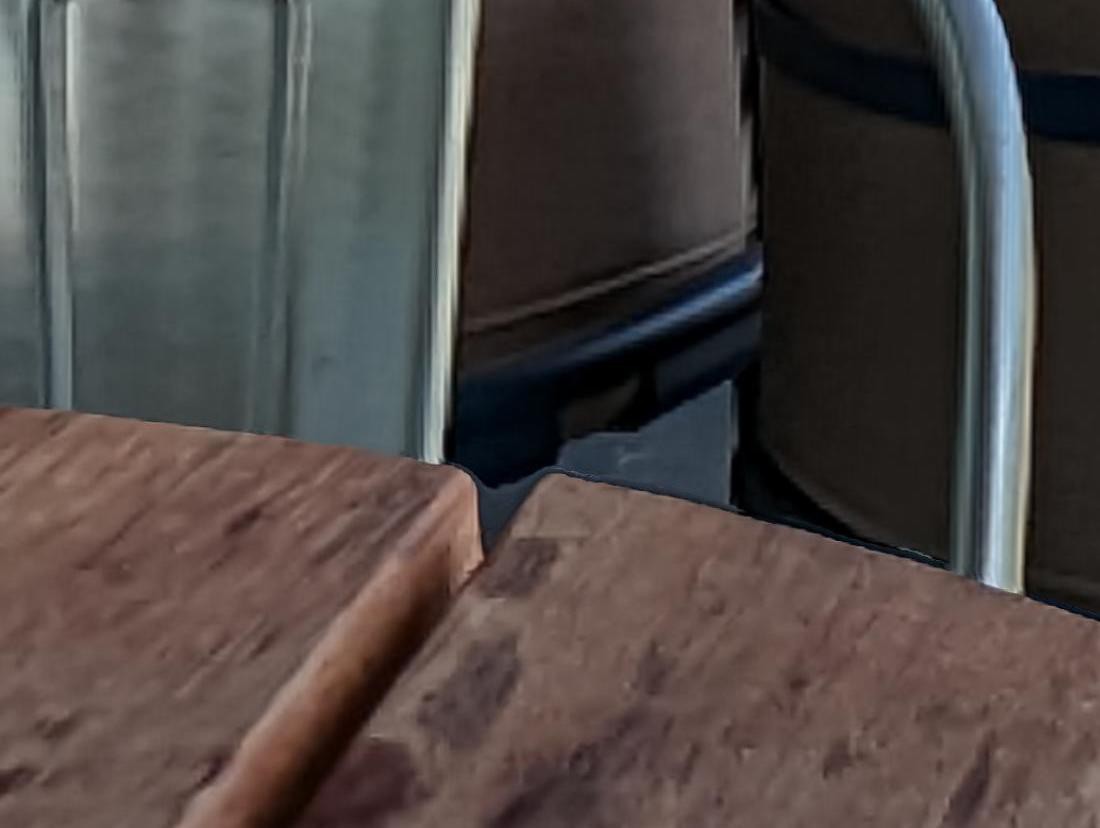}
    \mfigure{0.45}{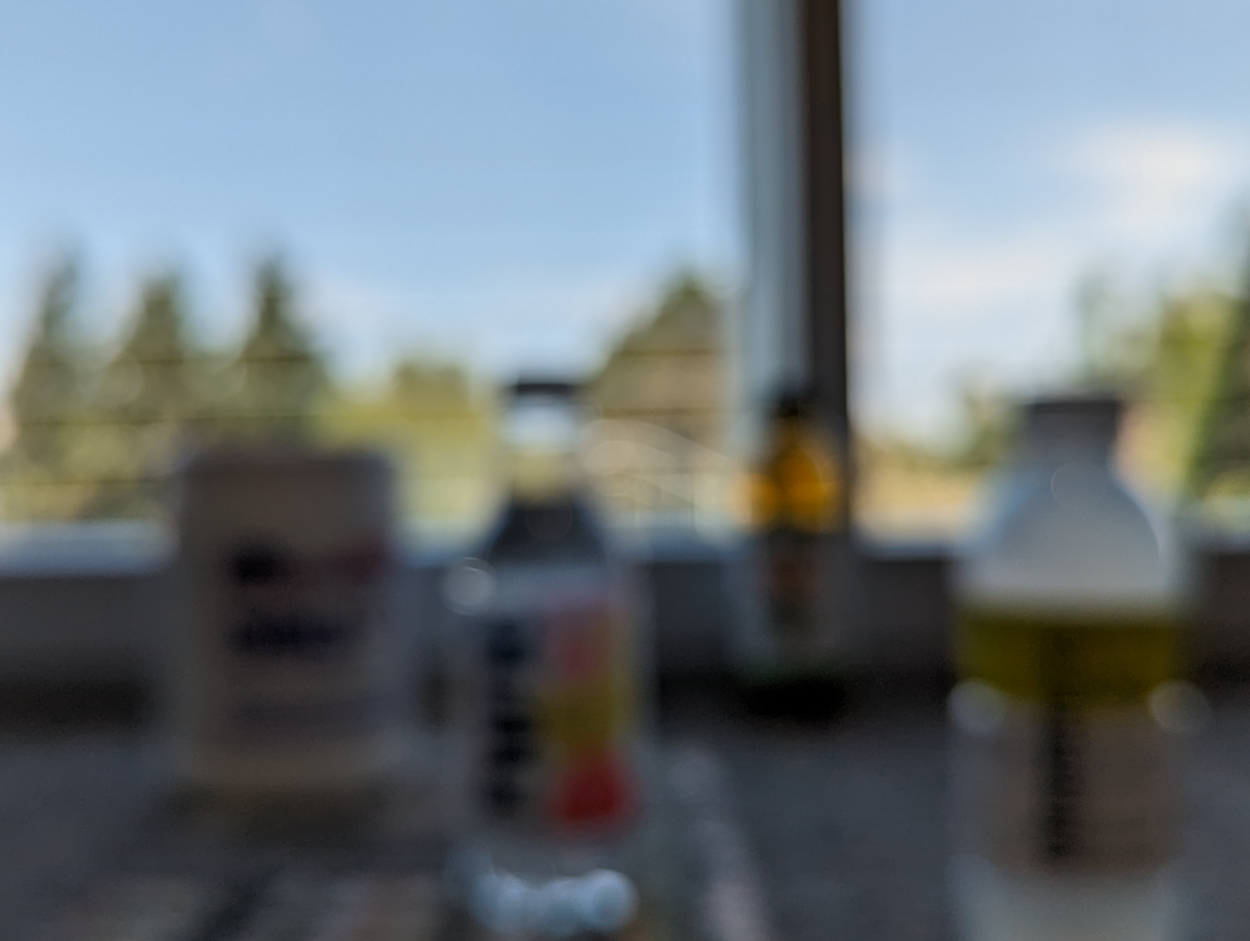} \hfill
    \mfigure{0.45}{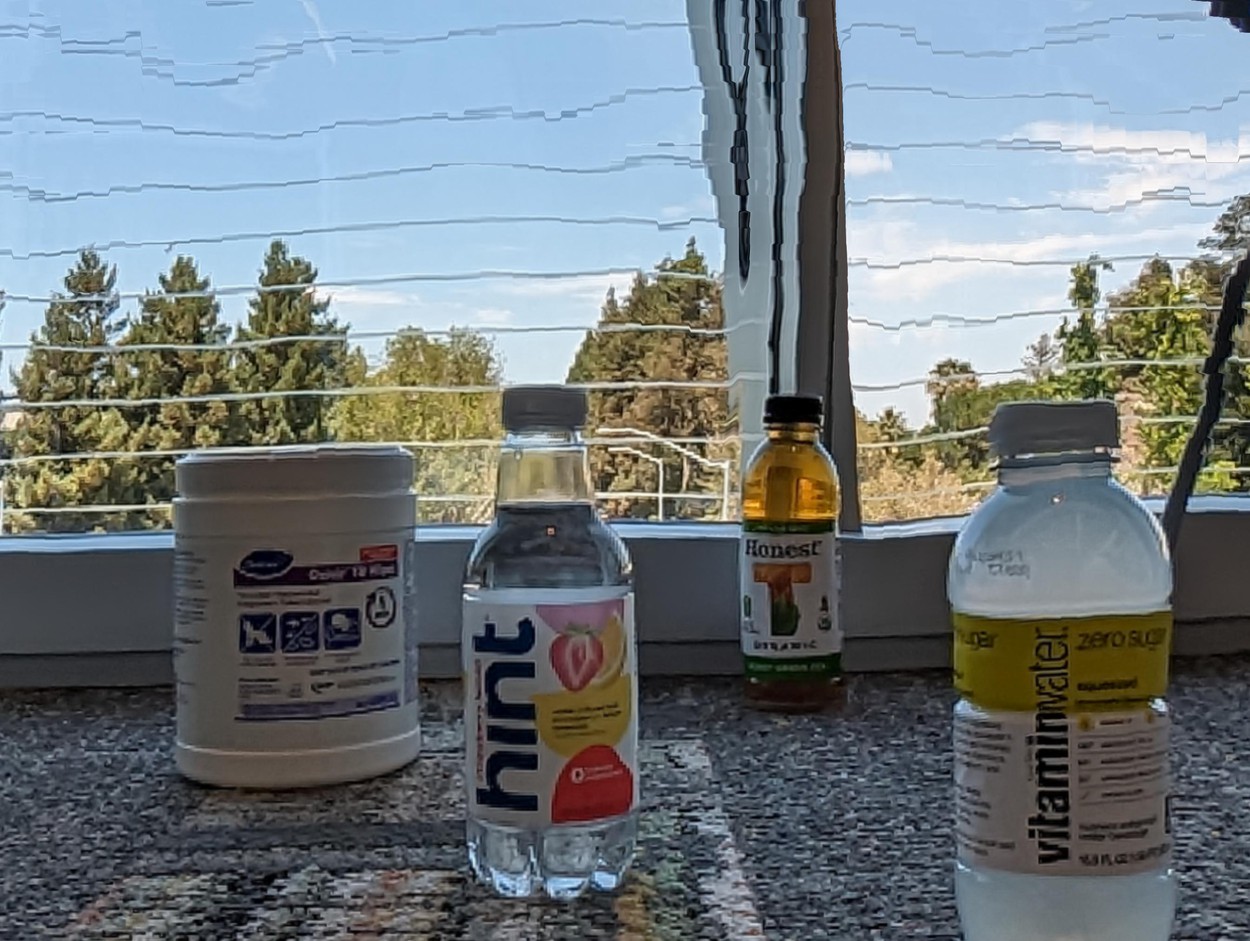} \hfill
    \mpage{0.45}{\textbf{W}} \hfill 
    \mpage{0.45}{Warped \textbf{UW}} \hfill 

    \caption{
    \textbf{Failure cases.} (\textit{Top}) In regions where \textbf{UW} is out of focus, as in the table shown above, our method would not be able to restore the sharp texture shown in the \textbf{W} image when focused on that object. (\textit{Bottom}) Aligning the \textbf{UW} using optical flow based alignment could suffer when \textbf{W} is severly blurred like shown on the warping artifacts on the window.}
    \label{fig:failure}
\end{figure}

\section{Limitations and Conclusion}
\label{sec:conclusion}
We present $\text{DC}^2$, a novel framework for defocus control with dual-camera consumer smartphones. We bypass the need for synthetic data and domain gap issues by training with real data captured with a smartphone device. We do so by re-framing the defocus control problem as refocus and designing a learning-based solution for the same. The key idea behind our method is to use \textbf{UW} input as an additional signal to control defocus of \textbf{W}. Naturally, a limitation then is the DoF of \textbf{UW} itself;  as objects outside of its DoF might not be sharper than in \textbf{W}. In general, our method  benefits from asymmetry in the \textbf{W} and \textbf{UW} camera configurations and likely won't perform as well in systems with identical cameras. Another limitation is our dependence on pre-existing optical flow and stereo depth algorithms which can suffer from severe artifacts with defocus blur (Figure \ref{fig:failure}). 
A promising avenues for future work includes utilizing additional cameras to jointly model both scene depth and defocus control.

\topic{Acknowledgment} We would like to thank Junlan Yang, Xiaotong Wu, Lun-Cheng Chu, Mauricio Delbracio, Yichang Shih, and Seang Chau for their support and fruitful discussions.

{\small
\bibliographystyle{ieee_fullname}
\bibliography{egbib}
}

 \newpage
 \setcounter{section}{0}
\renewcommand{\thesection}{\Alph{section}}

\section{Video Visualization}
One major advantage of our method is the fine-grained control we can have on the defocus control. As a result, we can directly simulate changing the focus distance and aperture smoothly just like if we had a DSLR camera with variable focal length and aperture. Please refer to the video provided in the supplementary materials.

\section{Detailed Architecture}
The model architecture consists of three primary modules: $\Phi_{ref}^W$ to refine \textbf{W}, $\Phi_{ref}^{UW}$ to refine \textbf{UW}, and a fusion model $\Phi_{fusion}$ to predict a blending mask to blend the refined outputs. Both $\Phi_{ref}^{W}$ and $\Phi_{ref}^{UW}$ use DRBNet architecture \cite{ruan2022learning} that utilize kernels prediction to refine the input. Each refinement module predicts intermediate outputs in a multi-scale setup that can be used to speed up training. Specifically, the model generates refined outputs at the following scales: 8x downsampled, 4x downsampled, 2x downsampled, and the original resolution. To be able to fuse all the multi-scale outputs, $\Phi_{fusion}$ consists of several Atrous Spatial Pyramid Pooling (ASPP) convolutions blocks \cite{chen2017rethinking} to predict blending mask for each scale. The ASPP blocks for each scale take the refined \textbf{W} and \textbf{UW} of the associated scale, as well as an upsampled blending mask from the previous ASPP block with a residual connection of the upsampled mask (except for the first ASPP block since it has no preceding blending mask). There are two hyperparameters associated with the blending block for each scale: (1) atrous rates for the atrous convolutions, and (2) the number of channels each intermediate step of atrous convolutions outputs. In table \ref{tab:dfnet_hyperparams}, we include a list of the hyperparameters for the blending block associated with each scale. \\
One issue with training the model in using cropped patches is that the blur kernel is spatially varying depending on the crop position. To resolve the ambiguity, we follow the solution proposed by Abuolaim \etal \cite{abuolaim2021learning} and concatenate a radial mask to the inputs of all modules where the pixel values of the mask are the distance from the original image center, normalized.

\begin{table}[ht]
\centering
\caption{
\textbf{Fusion model ($\Phi_{fusion}$) hyperparameters.} The hyeraprameters for the ASPP convolution blocks are the atrous rates for the atrous convolutions, and the channels each layer outputs. The number of atrous convolution layers is the size of the channel list. Note that the final output consists of two channels which correspond to the \textbf{W} and \textbf{UW} blending masks.
}
\begin{tabular}{lcc}
\toprule

\textbf{Blending Block} & atrous rates & channels  \\
\midrule
1/8x scale & 1,3,5 & 16, 32, 2 \\
1/4x scale & 1, 3, 6, 12 & 16, 32, 2 \\
1/2x scale & 1, 3, 6, 12, 15 & 16, 32, 2 \\
1x scale & 1,3,6, 12, 15, 18 & 16, 32, 32, 2 \\
\bottomrule
\end{tabular}
\vspace{-0.1in}
\label{tab:dfnet_hyperparams}
\end{table}

\begin{figure}[!ht]
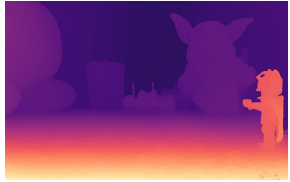

    \centering
    \mfigure{0.45}{iPhone_figures/iphone_disparity.jpg} \hfill

    \caption{\textbf{Disparity from iPhone}. Using portrait mode, we obtained Dual-camera disparity using an iPhone 14 Pro.}
    \label{fig:phone_disp}
\end{figure}

\section{Model Analysis}
\begin{figure*}[t!]
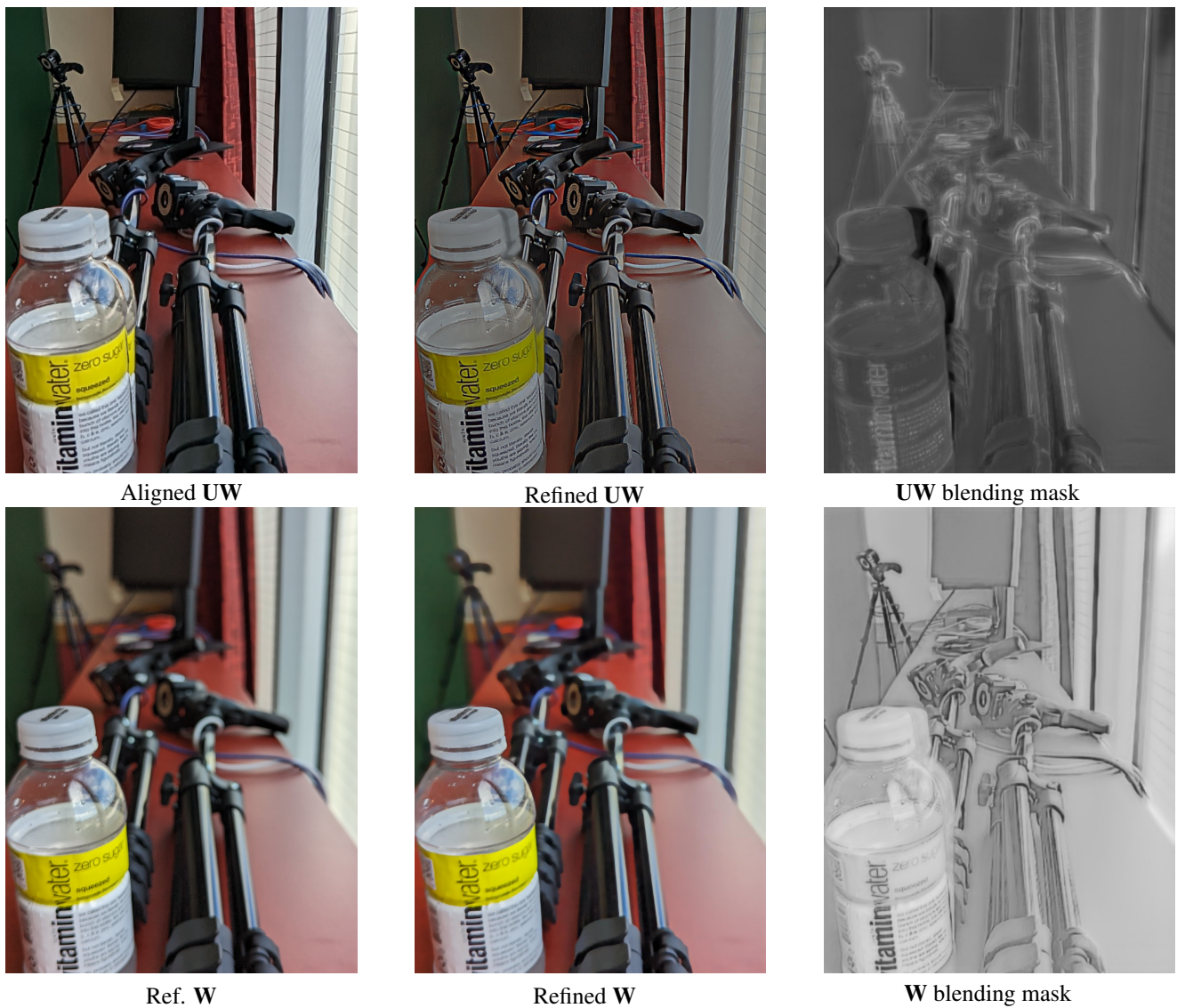

    \centering
    \mfigure{0.3}{supp_figures/tripod_src_uw.jpg} \hfill
    \mfigure{0.3}{supp_figures/tripoed_refined_uw.jpg} \hfill
    \mfigure{0.3}{supp_figures/tripod_uw_mask.jpg}
        \mpage{0.3}{Aligned \textbf{UW}} \hfill 
        \mpage{0.3}{Refined \textbf{UW}} \hfill
        \mpage{0.3}{\textbf{UW} blending mask} \hfill
        
    \mfigure{0.3}{supp_figures/tripod_src_w.png} \hfill
    \mfigure{0.3}{supp_figures/tripod_refined_w.jpg} \hfill
    \mfigure{0.3}{supp_figures/tripod_w_mask.jpg}
        \mpage{0.3}{Ref. \textbf{W}} \hfill 
        \mpage{0.3}{Refined \textbf{W}} \hfill
        \mpage{0.3}{\textbf{W} blending mask} \hfill

    \caption{
    \textbf{Intermediate results visualization.} Note that while the aligned \textbf{UW} suffers from an alignment artifcat around the bottle, the predicted masks take that into account by setting a low blending value for the occluded region in the \textbf{UW} mask and a higher value in the \textbf{W} mask.}
    \label{fig:analysis_tripod}
\end{figure*}

The primary motivation behind our architecture design is, depending on the target defocus map, the network can choose to deblur/blur parts of \textbf{W} and transfer sharper details from \textbf{UW} if necessary. Due to our model design, we can directly visualize the intermediate outputs to understand the model behavior. Specifically, we can visualize the refined \textbf{W} and \textbf{UW} which are the outputs of  $\Phi_{ref}^{W}$ and  $\Phi_{ref}^{UW}$, as well as the blending masks predicted by the fusion network $\Phi_{fusion}$. We visualize the intermediate outputs of our method on the task of all-in-focus deblurring in Figures \ref{fig:analysis_piano} and \ref{fig:analysis_tripod}. Note that in both examples, the mask associated with \textbf{UW} has large values around edges and regions with high frequencies, while the mask for \textbf{W} has higher values inlow frequency regions. This supports our hypothesis of having \textbf{UW} serve for high frequency details that could be blurry in \textbf{W}, while the \textbf{W} should be used as a reference to preserve the desired colors even in blurry regions. This behavior makes our method robust to color differences in \textbf{W} and \textbf{UW} just like we show in Figure \ref{fig:analysis_piano} where \textbf{UW} has incorrect white balance, and in Figure \ref{fig:analysis_tripod} we show how the model avoids relying on \textbf{UW} in occluded regions where artifacts may show up in the optical flow alignment. 

\begin{figure*}[t!]
    \centering
    \mfigure{0.3}{supp_figures/piano_src_uw.jpg} \hfill
    \mfigure{0.3}{supp_figures/piano_refined_uw.jpg} \hfill
    \mfigure{0.3}{supp_figures/piano_uw_mask.jpg}
        \mpage{0.3}{Aligned \textbf{UW}} \hfill 
        \mpage{0.3}{Refined \textbf{UW}} \hfill
        \mpage{0.3}{\textbf{UW} blending mask} \hfill
        
    \mfigure{0.3}{supp_figures/piano_src_w.png} \hfill
    \mfigure{0.3}{supp_figures/piano_refined_w.jpg} \hfill
    \mfigure{0.3}{supp_figures/piano_w_mask.jpg}
        \mpage{0.3}{Ref. \textbf{W}} \hfill 
        \mpage{0.3}{Refined \textbf{W}} \hfill
        \mpage{0.3}{\textbf{W} blending mask} \hfill

    \caption{
    \textbf{Intermediate results visualization.} Note that the whitebalance is off in \textbf{UW}, but the refinement module does not get affected by that since it primarily preserves the high frequencies in refined \textbf{UW}. In the refinement of \textbf{W}, we notice that the model deblurs the edges and preserves the low-frequency signals that can be blended with the details from \textbf{UW}}
    \label{fig:analysis_piano}
\end{figure*}

\section{Generalizing to Different Phone Setup}
\begin{figure*}[!th]
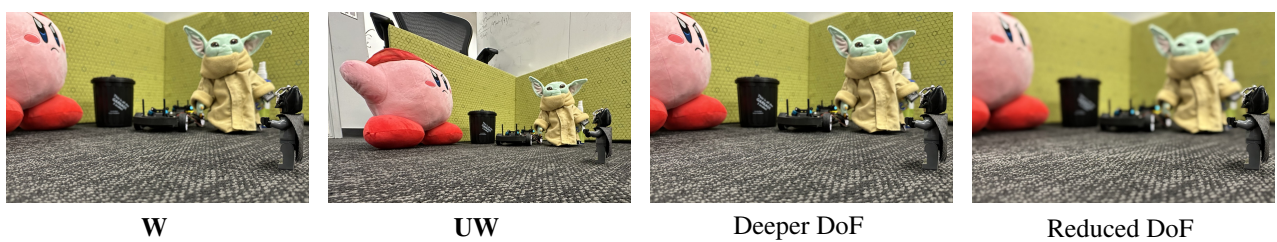

    \centering
    \mfigure{0.23}{iPhone_figures/w_toys.jpg} 
    \mfigure{0.23}{iPhone_figures/uw_toys.jpg}
    \mfigure{0.23}{iPhone_figures/deep_toys.jpg}
    \mfigure{0.23}{iPhone_figures/shallow_toys.jpg}
    \mpage{0.23}{\textbf{W}} \mpage{0.23}{\textbf{UW}} 
    \mpage{0.23}{Deeper DoF} \mpage{0.23}{Reduced DoF} 

    \caption{\textbf{Results on iPhone 14 Pro}. We ran our model on images from an iPhone 14 Pro, and show that it generalizes with blurring and deblurring despite not finetuning the model on any iPhone data.}
    \label{fig:iphone}
\end{figure*}

Our method requires only two cameras with different DoFs. This is widely available in modern smartphones since ultra-wide cameras tend to have a deeper DoF due to the small focal length compared to the wide and Telephoto cameras. Our approach that utilizes the defocus map is not specific to a particular device, but rather it can produce fairly good results for any smartphone with a similar \textbf{UW}+\textbf{W} dual-camera setup. To use data captured using an iPhone 14 Pro, we used the iPhone's portrait mode to obtain a disparity map (shown in Fig.~\ref{fig:phone_disp}), and warped the \textbf{UW} using an opitcal-flow based alignment. In Fig.~\ref{fig:iphone}, we show results of our model on data captured by iPhone 14 Pro~\textit{without any finetuning}.

\end{document}